\pdfoutput=1
\documentclass[11pt]{article}
\usepackage[preprint]{acl}
\usepackage{times}
\usepackage{latexsym}
\usepackage[T1]{fontenc}
\usepackage[utf8]{inputenc}
\usepackage{microtype}
\usepackage{inconsolata}
\usepackage{times}
\usepackage{latexsym}
\usepackage{soul}
\usepackage{amssymb}
\usepackage{amsmath}
\usepackage{booktabs}
\usepackage{bm}
\usepackage{cleveref}
\usepackage{graphicx}
\usepackage{tabularx}
\usepackage{soul}
\usepackage{tikz}
\usetikzlibrary{positioning, shadows.blur, fit, backgrounds, shapes.geometric}
\usepackage{xcolor}
\usepackage{todonotes}
\usepackage{multirow}
\usepackage{xpatch}
\usepackage{blindtext}
\usepackage{fdsymbol}
\usepackage{microtype}
\usepackage{hyperref}
\usepackage{adjustbox}
\usepackage{textcomp}
\usepackage{xparse}
\usepackage{enumitem}
\usepackage{makecell}
\usepackage{subcaption}
\usepackage{colortbl}
\usepackage{geometry}
\usepackage{algorithm}
\usepackage{authblk}
\usepackage{algpseudocode}
\usepackage{longtable}
\usepackage[indentfirst=false,rightmargin=0pt]{quoting}
\usepackage{stfloats} 

\usepackage{xparse}
\usepackage{highlight}

\definecolor{gold}{rgb}{0.83, 0.69, 0.22}
\NewDocumentCommand{\steeve}
{ mO{} }{\textcolor{gold}{\textsuperscript{\textit{Steeve}}\textsf{\textbf{\small[#1]}}}}

\title{NewsEdits 2.0: Learning the Intentions Behind Updating News}

\author[1]{\bf Alexander Spangher}
\author[2]{\bf Kung-Hsiang Huang}
\author[1]{\bf Hyundong Cho}
\author[1]{\bf Jonathan May}
\affil[1]{University of Southern California, Information Sciences Institute}
\affil[2]{Salesforce AI}
\affil[ ]{\texttt{spangher@usc.edu}}

\begin{document}
\maketitle
\begin{abstract}
As events progress, news articles often update with new information: if we are not cautious, we risk propagating outdated facts. In this work, we hypothesize that linguistic features indicate factual fluidity, and that we can \textit{predict which facts in a news article will update} using solely the text of a news article (i.e. not external resources like search engines). We test this hypothesis, first, by isolating fact-updates in large news revisions corpora \cite{spangher2022newsedits}. News articles may update for many reasons (e.g. factual, stylistic, narrative). We introduce the \textit{NewsEdits 2.0} taxonomy, an edit-intentions schema that separates fact updates from stylistic and narrative updates in news writing. We annotate over 9,200 pairs of sentence revisions and train high-scoring ensemble models to apply this schema. 
Then, taking a large dataset of silver-labeled pairs, we show that we can predict when facts will update in older article drafts with high precision. Finally, to demonstrate the usefulness of these findings, we construct a language model question asking (LLM-QA) abstention task. Inspired by \newcite{kasai2022realtime}, we wish the LLM to abstain from answering questions when information is likely to become outdated. Using our predictions, we show, LLM absention reaches \textit{near oracle levels of accuracy}. 
\end{abstract}

\section{Introduction}

News is the ``first rough draft of history'' \cite{croly1943new}. Its information is both valuable and fluid, prone to changes, updates, and corrections.
\begin{figure}[t]
    \centering
    \includegraphics[width=\linewidth]{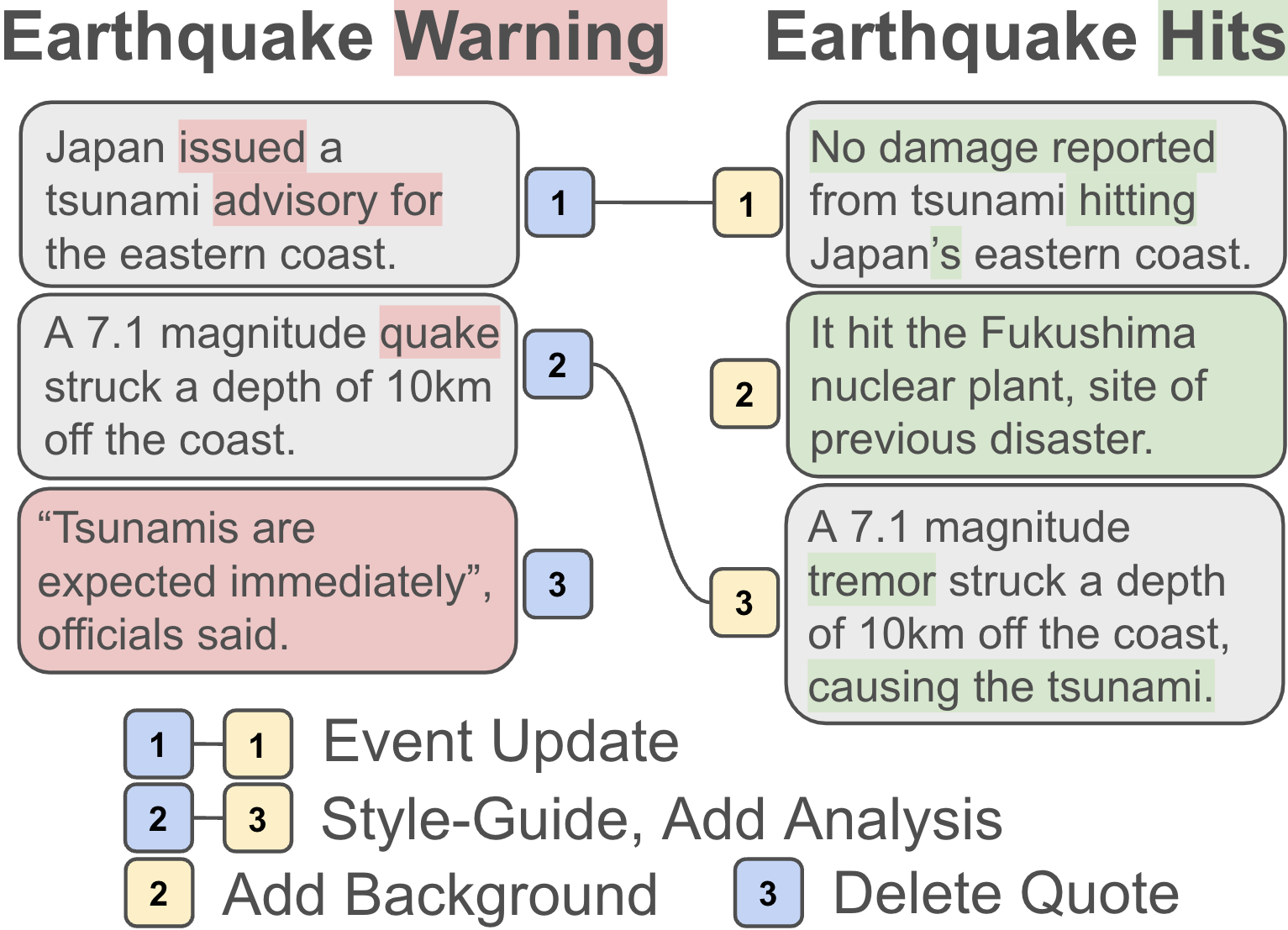}
    \caption{Updates can occur for many different reasons. 
    Shown here, we identify factual updates (e.g. ``Event Update'' between 1-1), stylistic updates (e.g. ``Style-Guide'' between 2-3) and narrative updates (e.g. ``Add Background'' for sentence addition 2). 
    }
    \label{fig:cover-figure}
\end{figure}
%
As shown in Figure \ref{fig:cover-figure}, the first sentence on the left
has a factual update, while the second
does not.
Intuitively, we might be able to predict this: an ``\textit{advisory}'' is not likely to indefinitely stay in effect, while details about the ``\textit{quake}'' are less likely to change. Indeed, if someone asks ``\textit{Q: Is an advisory still in place?}'', we might want to abstain from answering definitively. However, ``\textit{Q: How large was the quake?}'' can be answered directly. 

Recent work has recognized the importance of testing LLM-QA in dynamic settings \cite{jia2018tempquestions, liska2022cyprien}. \newcite{kasai2022realtime}'s RealTimeQA benchmark specifically measures LLM-QA performance for updating news documents. However, current approaches rely on search engines retrieving updated information\footnote{The latest entry of RealTimeQA was \textit{RAG + Google Custom Search}. \url{https://realtimeqa.github.io/}.}. This neglects potentially salient linguistic and common-sense information. As the example shown in Figure \ref{fig:cover-figure} demonstrates, cues exist that we, as humans, intuitively understand to signal fluidity. \textit{Can we learn these cues, and predict which facts in a news article will update? Can this help LLMs better abstain from answering questions they may not have updated information for?} 

\begin{figure*}
    \centering
    \includegraphics[width=\linewidth]{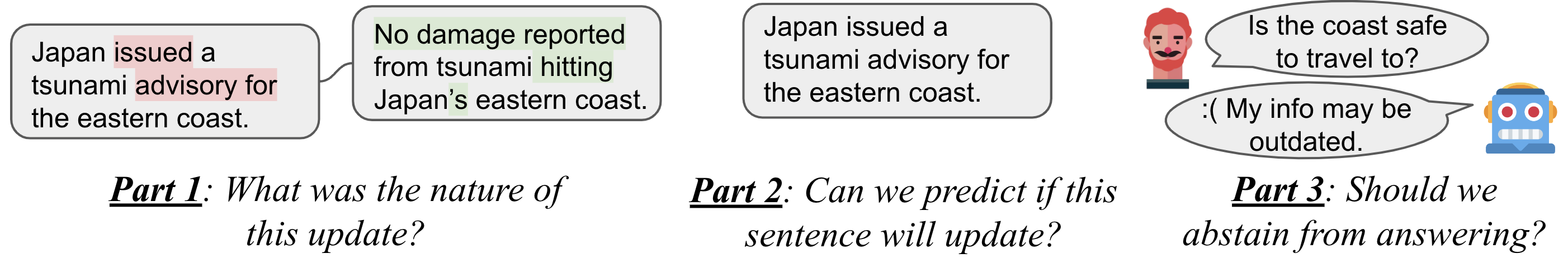}
    \caption{Overall paper flow. In \textbf{Part 1} of our paper, we develop an edits-intention scheme to describe news edits and train models to apply this schema to existing news revision corpora \cite{spangher2022newsedits}. In \textbf{Part 2}, we use these models to silver-label a large corpus and ask how well we can \textit{predict} whether a sentence will factually update. In \textbf{Part 3}, we show these predictions can be beneficial for increasing abstention rates during LLM-QA.}
    \label{fig:overall-flow}
\end{figure*}

We answer these questions in three steps, shown in Figure \ref{fig:overall-flow}. In \textbf{Part 1}, we start by studying update patterns in \textit{NewsEdits}, a large corpus of article revision histories \cite{spangher2022newsedits}. Articles update for many different reasons (e.g. factual, stylistic, etc.), and it is difficult to identify these reasons. So, we introduce \textit{NewsEdits 2.0}, a taxonomy of edit-intentions for journalistic edits (Figure \ref{fig:taxonomy-edits}), to help us do this. 
We hire professional journalists to annotate  9,200 pairs of sentence revisions across 507 article revision pairs with the \textit{NewsEdits 2.0} schema. We then train an ensemble model to tag pairs of revisions, with 75.1 Micro F1 and create a large silver-label corpus of revision pairs. 

Next, in \textbf{Part 2}, we use this silver-labeled corpus to predict which facts in articles might update. We find that models achieve a moderate macro-F1 of .58, overall, on a gold-labeled test set. \textit{\ul{Although these scores are noisy, we notice that our models are learning reasonable linguistic cues}}. We observe key linguistic patterns: the use of future-tense verbs, statistics and commonly updating events. We validate these cues with human measurement. Further, by focusing on the sentences our models predict are \textit{highly likely} to update, we notice a much higher precision of .74. Finally, in \textbf{Part 3}, we simulate a RealTimeQA-style case where an LLM using Retrieval Augmented Generation (RAG) retrieves an outdated document. Without our predictions, the LLM abstains wrongly more than it should. With them, the LLM achieves near-oracle level performance. 
In sum, our contributions are:
\begin{itemize}
    \item We introduce the \textit{NewsEdits 2.0} schema, with 4 coarse and 20 fine-grained categories, developed with professional journalists; train models to label these with 75.1 micro-F1; and release a large corpus of 4 million revision histories silver-labeled with edit intentions.
    \item We show that pretrained LLMs perform poorly at \textit{predicting which facts in the old versions articles will update}, indicating that this important capability is not emergent during pretraining. While fine-tuning helps performance, LLMs still lag humans.
    \item Finally, we show via a use-case, Question Answering with Outdated Documents, that a failure to address these shortcomings can result in decreased performance for leading LLMs. 
\end{itemize}

Finally, two subtle yet significant contributions of this work are (1) preprocessing improvements we introduce to improve the \textit{NewsEdits} corpus (e.g. improving sentence boundary detection); and (2) visualization tools to make revision histories more accessible to users. Because these advances are not relevant to the main ideas of our paper, we save a deeper discussion these for Appendix \ref{sct:technical_improvements}. 
Taken together, we hope that our work can increase utilization and understanding of news dynamics.



\section{Related Work}

Although most LLM Q\&A benchmarks assume that information is static, recent work has increasingly explored LLM performance in the presence of dynamic, updating information  \cite{jia2018tempquestions, liska2022cyprien}. This growing direction is concisely captured by \newcite{kasai2022realtime}'s statement: ``\textit{GPT-3 tends to return outdated answers when retrieved documents [are outdated]. Can [we] identify such unanswerable cases?}'' 

To our knowledge, the use of revision-histories to address this question, which we discuss in Section \ref{sct:use-case}, is novel. News updates are an especially crucial domain to study: (1) news is socially important \cite{cohen2011computational}; (2) LLMs are increasingly using news to better serve users \cite{hadero2023new}; (3) news is more likely to deal with updating events than other domains \cite{spangher2022newsedits}. Indeed, \newcite{kasai2022realtime}'s RealTimeQA benchmark is built entirely on news data.

Edit-intention schemas have been developed for other types of revision histories, like Wikipedia \cite{yang2017identifying}, and Student Learner Essays \cite{zhang2015annotation}. In these works, researchers categorize the intention of each edit using similar schemas to what we have developed. While building \textit{NewsEdits 2.0}, we were inspired by the schemas developed by prior work and they provided a starting point for our taxonomy. We added edit-categories that were more journalism specific, like ``Add Eye-witness Account'', and removed categories that were more specific to the aforementioned domains (Section \ref{sct:news-edits-schema},). The use-cases of these schemas has mainly focused on stylistic prediction tasks  (e.g. text simplification \cite{woodsend2011wikisimple} and grammatical error correction \cite{faruqui2018wikiatomicedits}) or tasks specific to these corpora (e.g. building models to assess the validity of a student's draft \cite{zhang2015annotation}, or counter vandalism on Wikipedia \cite{yang2017identifying}). We are the first, to our knowledge, to develop tasks centered on news articles (Section \ref{sct:predictive-modeling}) and to apply predictive analyses to fact-based edits. 

\begin{figure*}[ht!]
\centering
\begin{tikzpicture}[
    node distance=1cm and 0.5cm,
    title/.style={font=\bfseries\normalsize, fill=blue!50, text=white, align=center, minimum height=0.6cm, minimum width=4.6cm},
    item/.style={font=\small, align=center, text width=4.3cm, fill=blue!10},
    group/.style={draw, inner sep=1pt, rectangle},
]

\node[title] (factualTitle) {Factual Edit};
\node[item, below=0.1cm of factualTitle] (factual1) {Delete/Update/Add Eye-witness Account};
\node[item, below=0cm of factual1] (factual2) {Delete/Add/Update Event};
\node[item, below=0cm of factual2] (factual3) {Delete/Add/Update Source-Doc.};
\node[item, below=0cm of factual3] (factual4) {Correction};
\node[item, below=0cm of factual4] (factual5) {Delete/Add/Update Quote};
\node[item, below=0cm of factual5] (factual6) {Additional Sourcing (Other)};
\node[item, below=0cm of factual6] (factual7) {Additional Information (Other)};
\node[group, fit=(factualTitle) (factual7)] {};

\node[title, right=0.25cm of factualTitle] (styleTitle) {Style edit};
\node[item, below=0.1cm of styleTitle] (style1) {Simplification};
\node[item, below=0cm of style1] (style2) {Emphasize/De-emphasize Importance};
\node[item, below=0cm of style2] (style3) {Define term};
\node[item, below=0cm of style3] (style4) {Style-Guide Adherence};
\node[item, below=0cm of style4] (style5) {Syntax Correction};
\node[item, below=0cm of style5] (style6) {Tonal Edits};
\node[item, below=0cm of style6] (style7) {Sensitivity Consideration};
\node[group, fit=(styleTitle) (style7)] {};

\node[title, right=0.25cm of styleTitle] (backgroundTitle) {Narrative/Contextual};
\node[item, below=0.1cm of backgroundTitle] (background1) {Delete/Add/Update Analysis};
\node[item, below=0cm of background1] (background2) {Delete/Add/Update Background};
\node[item, below=0cm of background2] (background3) {Delete/Add/Update Anecdote};
\node[group, fit=(backgroundTitle) (background3)] {};

\node[title, below=0.19cm of background3] (otherTitle) {Other};
\node[item, below=0.1cm of otherTitle] (other1) {Incorrect Link};
\node[item, below=0cm of other1] (other2) {Unchanged};
\node[item, below=0cm of other2] (other3) {Other/None};
\node[group, fit=(otherTitle) (other3)] {};

\end{tikzpicture}
\caption{\textit{NewsEdits 2.0}: Edit-Intentions Schema categories and their subcategories. In this work, we focus mainly on the \textit{Factual Edit} category. See Appendix \ref{apx:annotation_guidelinnes} for definitions for all categories.}
\label{fig:taxonomy-edits}
\end{figure*}

\section{Part 1: Learning Edit Intentions in Revision Histories}
\label{sct:newsedits-schema-and-model}

News articles update for different reasons, especially during breaking news cycles where facts and events update quickly \cite{saltzis2012breaking}. In this section, we introduce the edit-intentions schema we use for \textit{NewsEdits 2.0}, our annotation, and our models to label edit-pairs. This lays groundwork for Section \ref{sct:predictive-modeling}, where we will predict when facts change. 

We wish to identify categories of edits, in order to enable different investigations into these different update patterns. In other words, we describe the following update model:

\begin{equation}
    p( l | s_i, s'_j, D, D')
    \label{eq:label_model}
\end{equation}

\noindent where $l$ is an \textit{intention} (e.g. a ``Correction'' needs to be made), $D$ and $D'$ represent the older and newer versions of a news article, respectively, and $s_i$ and $s'_j$ are individual sentences where the update occurred. $i, j$ are sentence indices, ranging from $i \in \{1, ... n\}$, $j \in \{1, ... m\}$ (where $n, m$ are the number of sentences in $D$, $D'$).  

\subsection{Edit Intentions Schema}
\label{sct:news-edits-schema}

 We work with two professional journalists and one copy editor\footnote{Collectively, these collaborators have over 50 years of experience in major newsrooms.} to develop an intentions schema. Building off work by \newcite{zhang2015annotation} and \newcite{yang2017identifying}, we start by examining 50 revision-pairs sampled from \textit{NewsEdits}. We developed our schema through 4 rounds of conferencing: tagging examples finding edge-cases and discussing whether to add or collapse schema categories. Figure \ref{fig:taxonomy-edits} shows our schema, which we organize into coarse and fine-grained labels. 
We incorporate existing theories of news semantics into our schema. For instance, ``Event Updates'' incorporates definitions of ``events'' \cite{doddington2004automatic}, while ``Add Background'' incorporates theories of news discourse \cite{van1998newsdiscourse}. ``Add Quote'' incorporates definitions from informational source detection \cite{spangher2023identifying} and ``Add Anecdote'' incorporates definitions from editorial analysis \cite{alkhatib2016b}. See Appendix \ref{app:additional_schema_definitions} for a deeper discussion of the theoretical schemas that inform the NewsEdits 2.0 schema. Finally, ``Incorrect Link'' is an attempt to correct sentence pairs that were erroneously (un)linked in \textit{NewsEdits}.

\subsection{Schema Annotation}
\label{sct:annotation}

We build an interface for annotators to provide intention labels for news article sentence pairs (see Appendix \ref{app:annotation_interface}). Annotators are shown definitions for each fine-grained intention and the articles to tag; they are instructed to tag each sentence. 
To recruit annotators, we posted on two list-serves for journalism industry professionals\footnote{The Association of Copy Editors (ACES) \url{https://aceseditors.org/} and National Institute for Computer-Assisted Reporting (NICAR) \url{https://www.ire.org/hire-ire/data-analysis/}.}. We train our annotators until they are all tagging with $\kappa > .6$ agreement, compared with a gold-set of 50 article revision-pairs that we annotated, described previously (Section \ref{sct:news-edits-schema}). See Appendix for more details.

\subsection{Edit Intentions Modeling}
\label{sct:revision-pair-modeling}

\begin{table*}[t]
    \small
    \centering
    \begin{tabular}{lrrrrrrrr}
        \toprule
        & \multicolumn{2}{c}{\textbf{All}} & \multicolumn{2}{c}{\textbf{Fact}}& \multicolumn{2}{c}{\textbf{Style}} & \multicolumn{2}{c}{\textbf{Narrative}} \\
        \cmidrule(lr){2-3} \cmidrule(lr){4-5} \cmidrule(lr){6-7} \cmidrule(lr){8-9}
Features &  Macro &  Micro &  Macro &  Micro & Macro &  Micro & Macro & Micro \\
\midrule
Baseline, \textit{fine-grained}                        &      \gca{45.8} &      \gcb{73.6} &           \gcc{32.0} &           \gce{47.2} &            \gcf{58.6} &            \gcg{39.9} &                \gch{52.0} &                \gci{39.9} \\
\midrule
+ NLI                           &      \gca{48.6} &      \gcb{74.1} &           \gcc{45.7} &           \gce{50.4} &            \gcf{55.2} &            \gcg{38.7} &                \gch{43.6} &                \gci{38.7} \\
+ Event                         &      \gca{46.7} &      \gcb{74.1} &           \gcc{39.0} &           \gce{49.0} &            \gcf{59.3} &            \gcg{41.4} &                \gch{41.7} &                \gci{41.4} \\
+ Quote                         &      \gca{46.3} &      \gcb{72.8} &           \gcc{49.8} &           \gce{54.7} &            \gcf{31.9} &            \gcg{28.0} &                \gch{42.4} &                \gci{28.0} \\
+ Collapsed Quote               &      \textbf{\gca{51.2}} &      \gcb{73.9} &           \gcc{38.7} &           \gce{47.6} &            \gcf{58.3} &            \gcg{39.4} &                \gch{51.4} &                \gci{39.4} \\
+ Discourse                     &      \gca{45.8} &      \gcb{75.1} &           \gcc{37.7} &           \gce{49.6} &            \gcf{63.8} &            \gcg{44.6} &                \gch{43.2} &                \gci{44.6} \\
+ Argumentation                 &      \gca{48.9} &      \gcb{73.6} &           \gcc{37.1} &           \gce{47.9} &            \gcf{57.1} &            \gcg{37.7} &                \gch{53.5} &                \gci{37.7} \\
\midrule
+ Discourse \& Event             &      \gca{46.3} &      \gcb{74.3} &           \gcc{38.9} &           \gce{49.9} &            \gcf{62.1} &            \gcg{42.2} &                \gch{42.4} &                \gci{42.2} \\
+ Discourse \& Argumentation     &      \gca{47.8} &      \gcb{74.1} &           \gcc{56.8} &           \gce{50.5} &            \gcf{31.4} &            \gcg{32.2} &                \gch{41.1} &                \gci{32.2} \\
+ Argumentation \& Event         &      \gca{50.0} &      \gcb{75.1} &           \gcc{38.0} &           \gce{48.6} &            \gcf{46.4} &            \gcg{44.9} &                \gch{58.5} &                \gci{44.9} \\
+ Quote \& Discourse             &      \textbf{\gca{51.2}} &      \gcb{72.2} &           \gcc{40.5} &           \gce{45.3} &            \gcf{62.8} &            \gcg{43.0} &                \gch{48.7} &                \gci{43.0} \\
+ Collapsed Quote \& Discourse   &      \gca{49.6} &      \gcb{73.9} &           \gcc{45.6} &           \gce{49.4} &            \gcf{58.9} &            \gcg{39.1} &                \gch{47.9} &                \gci{39.1} \\
+ Collapsed Quote \& NLI         &      \gca{45.4} &      \gcb{72.8} &           \gcc{41.9} &           \gce{50.4} &            \gcf{46.7} &            \gcg{31.2} &                \gch{39.3} &                \gci{31.2} \\
\midrule
+ Collapsed Quote \& NLI \& Event &      \gca{49.0} &      \gcb{73.8} &           \gcc{44.9} &           \gce{48.9} &            \gcf{57.4} &            \gcg{37.0} &                \gch{44.0} &                \gci{37.0} \\
\midrule
+ All                           &      \gca{47.2} &      \gcb{73.6} &           \gcc{40.0} &           \gce{49.7} &            \gcf{58.6} &            \gcg{36.0} &                \gch{43.5} &                \gci{36.0} \\
\midrule 
Baseline, \textit{coarse-grained}   &   49.4 & 56.7 & \multicolumn{2}{c}{46.6}  & \multicolumn{2}{c}{65.1} & \multicolumn{2}{c}{10.4}\\
+ Discourse \& Arg. (Best model, Fact)         &   65.4 & 70.7 & \multicolumn{2}{c}{59.4} & \multicolumn{2}{c}{66.2} & \multicolumn{2}{c}{49.2}\\
\bottomrule
        \end{tabular}
    \caption{Various F1 scores (\%) on our test set of the fine-tuned LED model with different combinations of features. Fact/Style/Narrative F1 scores are computed on instances that contain the corresponding labels, whereas All F1 scores are derived from all instances. \looseness=-1}
    \label{tab:led_results}
\end{table*}

Now, we are ready to classify edit intentions between sentences in article revisions. 
Edit intentions are labeled on the sentence-level, and each sentence addition, deletion or update has potentially multiple intention-labels. Document-level context is important: as shown in Figure \ref{fig:cover-figure}, understanding that Sentence 2, right, adds background (``\textit{It hit the Fukushima plant, site of previous disaster.}'') 
is aided by the surrounding sentences contextualizing that a major event had just occurred.  So, we wish to construct models that can produce flexible outputs and reason about potentially lengthy inputs.

Generative models have recently been shown to outperform classification-based models in document understanding tasks \cite{li-etal-2021-document, huang-etal-2021-document}. Inspired by this, we develop a sequence-to-sequence framework using LongFormer \footnote{\url{https://huggingface.co/allenai/led-base-16384}} \cite{beltagy2020longformer} to predict the intent behind each edit. Specifically, our model processes the input $x = [s_i || s’_j || D || D']$. $s_i$ or $s'_j$ can also be $\varnothing$, which corresponds to the other sentence being a addition/deletion. The decoding target $y_{i, j} = [l_1 || … || l_k]$ is a concatenation of $\geq 0$ intention labels $1_1, ..., 1_k$ annotated for the pair $s_i, s'_j$. 

\paragraph{Experimental Variants} 

As discussed in Section \ref{sct:news-edits-schema}, we developed our schema to bring together different theories of news semantics. So, we hypothesize that incorporating insights from these theories into our modeling -- specifically, by utilizing labels from trained models in these domains -- might improve our performance. We run models from the following papers over our dataset: \textit{Discourse} \cite{spangher2021multitask}, \textit{Quote-Type Labeling} \cite{spangher2023identifying}, \textit{Event Detection} \cite{hsu2021degree}, \textit{Textual Entailment} \cite{nie-etal-2020-adversarial} and \textit{Argumentation} \cite{alkhatib2016b}. Labels generated from these models, denoted as $f_{s_i}$ and $f_{s’_j}$, are appended to the model input $x = [s_i || s’_j || D || D’ || f_{s_i} || f_{s’_j}]$. 

\paragraph{Edit-Intention Taggin Model Performance}

As shown in Table \ref{tab:led_results}, our baseline tagging models that solely use article features score 45.8 Macro F1 and 73.6 Micro F1, respectively. These scores are moderate-to-low. The category we are most interested in, Factual updates, scores at 32 Macro-F1 (derived from macro-averaging the fine-grained categories).
However, incorporating additional features increases overall Macro and Micro F1 by $5.5$ and $1.5$ points, respectively, in the \textit{Quotes \& Discourse} trial. And for Factual updates, additional features increase Macro and Micro F1 accuracy by $17.8$ and $7.5$ points, respectively. While low-to-moderate scores are not ideal, this likely reflects the noisy nature of our problem. \textit{We hope in future work to assess an upperbound on these scores.} For details and schema definitions, see \Cref{apx:led_details}.


\subsection{Exploratory Insights}
\label{sct:eda}

\paragraph{Different edit-intentions distribute differently across different edit types (Add, Deletion, Update).} 

We run the models trained in the last section over the entire \textit{NewsEdits} corpus to generate silver-labels on all edit pairs. We present an exploratory analysis of these silver labels, with more material shown in the appendix. Table \ref{tab:syntactic} shows the correlation between syntactic edit categories (defined by \cite{spangher2022newsedits}) and our semantic categories. As can be seen, categories like Addition have far more Narrative and Factual updates than Stylistic updates; Stylistic updates, on the other hand, are far more likely to occur between sentences. This is logical; Stylistic updates are likely smaller, local updates, while Narrative and Factual updates might include more rewriting.

\paragraph{Different edit-intentions distribute differently across different kinds of news (e.g. Business, Politics).} 

Next, we explore if certain \textit{kinds of articles} are more likely to have certain \textit{kinds of edits}. We start by looking at broad news categories, shown in Table \ref{tab:topic_classification_2}, obtained from classifier we train on CNN News Groups dataset\footnote{\url{https://www.kaggle.com/code/faressayah/20-news-groups-classification-prediction-cnns}}. ``Politics'' and ``Sports'' coverage are observed to have the highest level of Factual updates, relative to other categories, while Stylistic updates are prevalent in ``Health'' and ``Entertainment'' pieces. Although we focus on Factual updates for the rest of the paper, we believe that there are many fruitful directions of future work examining other categories of updates. For instance, stylistic edits made in ``Health'' news might reach more readers -- understanding these patterns might be crucial during times of crisis. We include additional exploration in Appendix \ref{app:eda}.  

\begin{table}
\small
\centering
\begin{tabular}{lrrr}
\toprule
 & \textbf{Narrative} & \textbf{Fact} & \textbf{Style} \\
\midrule
Addition & 840329 & 358900 & 104 \\
Deletion & 330039 & 21671 & 6088 \\
edit & 411292 & 102499 & 644243 \\
\bottomrule
\end{tabular}
\caption{Counts of coarse-grained semantic edit types, broken out by syntactic categories (for fine-grained counts, see Appendix).}
\label{tab:syntactic}
\end{table}

\begin{table}[t]
    \centering
    \small
    \begin{tabular}{lrrr}
    \toprule
 & Fact & Style & Narrative \\
 \midrule
Business & {\cellcolor[HTML]{FFF7FB}} \color[HTML]{000000} 1.6 & {\cellcolor[HTML]{4897C4}} \color[HTML]{F1F1F1} 62.0 & {\cellcolor[HTML]{4A98C5}} \color[HTML]{F1F1F1} 36.4 \\
Entertainment & {\cellcolor[HTML]{9EBAD9}} \color[HTML]{000000} 3.3 & {\cellcolor[HTML]{023858}} \color[HTML]{F1F1F1} 65.5 & {\cellcolor[HTML]{FFF7FB}} \color[HTML]{000000} 31.1 \\
Health & {\cellcolor[HTML]{EEE8F3}} \color[HTML]{000000} 2.1 & {\cellcolor[HTML]{80AED2}} \color[HTML]{F1F1F1} 61.0 & {\cellcolor[HTML]{308CBE}} \color[HTML]{F1F1F1} 36.9 \\
News & {\cellcolor[HTML]{C6CCE3}} \color[HTML]{000000} 2.8 & {\cellcolor[HTML]{FFF7FB}} \color[HTML]{000000} 57.0 & {\cellcolor[HTML]{023858}} \color[HTML]{F1F1F1} 40.2 \\
Politics & {\cellcolor[HTML]{023858}} \color[HTML]{F1F1F1} 5.9 & {\cellcolor[HTML]{F1EBF4}} \color[HTML]{000000} 57.8 & {\cellcolor[HTML]{509AC6}} \color[HTML]{F1F1F1} 36.3 \\
Sport & {\cellcolor[HTML]{8BB2D4}} \color[HTML]{000000} 3.5 & {\cellcolor[HTML]{C9CEE4}} \color[HTML]{000000} 59.3 & {\cellcolor[HTML]{2484BA}} \color[HTML]{F1F1F1} 37.2 \\
\bottomrule
\end{tabular}
    \caption{Distribution over update-types, across CNN section classifications.}
    \label{tab:topic_classification_2}
\end{table}

\section{Part 2: Predicting Factual Updates}
\label{sct:predictive-modeling}

In Section \ref{sct:newsedits-schema-and-model}, we learned high-scoring models to categorize edit pairs (Equation \ref{eq:label_model}). Now, we wish to leverage these to learn a predictive function:

\begin{equation}
    p(l = \text{Factual-Update} | s_i, D)
    \label{eq:prediction_task}
\end{equation}

Where $s_i$ and $D$ are the \textit{older} half of a revision pair. Eq \ref{eq:prediction_task} seeks to predict how $D$ \textit{might} change. 

The problem statement builds off of a line of inquiry introduced in \newcite{spangher2022newsedits}. Authors introduced tasks aimed at predicting news article developments across time. They tried to predict whether a ``sentence will be \textit{Added} to, \textit{Deleted} from, or \textit{Updated} in'' an older draft, to induce reasoning about article changes. 
However, authors stopped at this ``syntactic'' analysis. Here, we build off of this mode of inquiry: with the semantic understanding of edits introduced in the prior section, we try to predict \textit{how} information will change.

\subsection{Factual Edit Prediction Dataset}
\label{sec:predictive_modeling_data}

To construct our task dataset, we sample revision pairs with a non-negligible amount of updates. We sample a set of 500,000 articles from \textit{NewsEdits} that have $>10\%$ sentences added and $>5\%$ deleted. \textit{\ul{We acnkowledge that this introduces bias into our dataset}}, as we focus solely on a subsection of data we \textit{know} will update. However we build off \newcite{spangher2022newsedits}'s broader analysis of syntactic edits patterns, where they found that these kinds of articles could be predicted with reasonable accuracy. We reason that our construction makes it more likely that we are focusing on factual updates that have more significant impact on the article (as they require more substantial rewrites.)

Then, we use the best-performing edit-intentions model, in Section \ref{sct:revision-pair-modeling}, to produce silver labels. We assign labels $l$ using both versions of a revision pair (Equation \ref{eq:label_model}); then we discard $D'$, $s'_j$ and try to predict $l$ using \textit{just} $D, s_i$ (Equation \ref{eq:prediction_task}).

\begin{table*}[t]
    \small
    \centering
    \begin{tabular}{llrrrr}
        \toprule

        \textbf{Model} & \textbf{Features} & \textbf{Fact F1}  & \textbf{Not Fact F1} & \textbf{Macro F1 } &  \textbf{Micro F1}  \\ 
                
        \midrule 
        \multirow{3}{*}{GPT-3.5} 
        & Sentence-Only  & 11.3 & 79.1 & 30.4 & 74.2 \\ 
        & Direct Context & 3.4 & 91.8 & 32.2 & 85.2 \\ 
        & Full Article & 7.9 & 91.1 & 49.8 & 85.4 \\ 

        \midrule 
        \multirow{3}{*}{GPT-4} 
        & Sentence-Only  & 11.1 & 66.3 & 38.9 & 62.4  \\
        & Direct Context & 14.8 & 88.8 & 52.7 & 84.1 \\ 
        & Full Article & 15.4 & 90.6 & 53.2 & 84.9 \\ 

        \midrule 


        \multirow{3}{*}{FT Longformer} 
        & Sentence-Only  & 21.2 & 92.3 & 57.4 & 87.0 \\ 
        & Direct Context & 22.3 & 93.0 & 87.8 & 87.4  \\ 
        & Full Article & 25.4 & 91.4 & 58.0  & 86.4 \\
        \midrule
        Human Performance & Sentence-Only & 41.2 & 75.3 & 58.6 & 69.2 \\
      \bottomrule
        \end{tabular}
    \caption{How well can models predict if a sentence will have a fact update, or not? We test GPT3.5 and GPT4. Individual, macro and micro F1 scores (\%) on the golden test set for various evaluated models. \looseness=-1}
    \label{tab:predictive_modeling_results}
\end{table*}

\subsection{Predicting Factual Edits}

For training and development, we chronologically split our dataset into train/development sets with 80/20 ratios. The earliest 80\% is our training set, the next 20\% for development, etc. To keep cost reasonable, we sample 16,000 sentences for the training set and 2,000 for the development set. \textit{\ul{We test all approaches on the same gold-labeled documents $D_{test}^{gold}$, which were part of our gold-annotated test set}} (Section \ref{sct:annotation}). In early experiments, we noticed that many fine-grained labels were too infrequent to model well, so we switched to predicting coarse-grained labels. 
We balance the training dataset to have an equal number of classes.

\paragraph{Factual Edit Prediction Experiments}

We test different variants of Equation \ref{eq:prediction_task} to provide different degrees of article context to the model. This helps us understand how much local vs. global article features predict Factual Updates.

(1) \ul{Sentence-Only}, $p(l | s_i)$; 

(2) \ul{Direct Context}, $p(l | s_{i-1}, s_i, s_{i+1})$

(3) \ul{Full Article}, $p(l | s_i, D)$.

For each variant we test zero-shot (i.e. prompted \texttt{gpt-3.5-turbo} and \texttt{gpt-4}); and fine-tuning approaches (i.e. longformer models
)\footnote{The longformer is trained with the same approach as the silver-label prediction step from Section \ref{sct:revision-pair-modeling}
In early trials, we try different variations on these experiments, like restricting the dataset to different subsets based on topic, like ``Disaster'' or ``Safety''. These topic categories, as shown in Section \ref{sct:eda}, are more fact-heavy. However, we find negligible impact on F1-score.}. 
%

Results are shown in Table \ref{tab:predictive_modeling_results}. Performance is moderate-to-low for detecting factual updates. However, we do observe performance increases from fine-tuning the longformer model, so to some degree this task is learnable. We recruit a former journalist, with 4 years of experience in major newsrooms, to predict labels for this task, in order to provide a human upper bound to Equation \ref{eq:prediction_task}. The journalist observes the training data, and then scores the test set. At 41.2 F1-score, the journalist sets a moderately higher upper bound.

\begin{table}[t]
    \centering
    \small
    \begin{tabular}{lrrr}
    \toprule
    \textbf{Sent. Contains:} & Fact U. & $\overline{\text{Fact U.}}$ & $\Delta$ \\
    \midrule
Recent Event & {\cellcolor[HTML]{AAAAFF}} \color[HTML]{000000} 50\% & {\cellcolor[HTML]{F2F2FF}} \color[HTML]{000000} 8\% & {\cellcolor[HTML]{76FF76}} \color[HTML]{000000} 42\% \\
Developing Event & {\cellcolor[HTML]{CBCBFF}} \color[HTML]{000000} 30\% & {\cellcolor[HTML]{FFFFFF}} \color[HTML]{000000} 0\% & {\cellcolor[HTML]{A0FFA0}} \color[HTML]{000000} 30\% \\
Statistic & {\cellcolor[HTML]{D1D1FF}} \color[HTML]{000000} 28\% & {\cellcolor[HTML]{F2F2FF}} \color[HTML]{000000} 8\% & {\cellcolor[HTML]{C4FFC4}} \color[HTML]{000000} 19\% \\
Info. request & {\cellcolor[HTML]{EAEAFF}} \color[HTML]{000000} 12\% & {\cellcolor[HTML]{FFFFFF}} \color[HTML]{000000} 0\% & {\cellcolor[HTML]{D8FFD8}} \color[HTML]{000000} 12\% \\
Historical Event & {\cellcolor[HTML]{FFFFFF}} \color[HTML]{000000} 0\% & {\cellcolor[HTML]{E3E3FF}} \color[HTML]{000000} 17\% & {\cellcolor[HTML]{FFC9C9}} \color[HTML]{000000} -17\% \\
Opinion/Analysis & {\cellcolor[HTML]{FCFCFF}} \color[HTML]{000000} 2\% & {\cellcolor[HTML]{BFBFFF}} \color[HTML]{000000} 39\% & {\cellcolor[HTML]{FF8B8B}} \color[HTML]{000000} -36\% \\
Description & {\cellcolor[HTML]{F0F0FF}} \color[HTML]{000000} 10\% & {\cellcolor[HTML]{AAAAFF}} \color[HTML]{000000} 50\% & {\cellcolor[HTML]{FF8181}} \color[HTML]{F1F1F1} -40\% \\
    \bottomrule
    \end{tabular}
    \caption{\textbf{Linguistic Cues characterizing Factual Updates}: Manual annotations of characteristics in $D_{test}^{gold}$ sentences that either Factually Update, or not. We show the \% of sentences containing these characteristics, ordered by those most salient for Factual Updates.}
    \label{tab:linguistic_qualities}
\end{table}

\begin{table}[t]
    \centering
    \begin{tabularx}{\linewidth}{>{\hangindent=2em}X}
    \toprule
    Sentences with $\uparrow p(l | s_i, D)$ \\
    \midrule
        There are no immediate reports of casualties. \\
        His trial has not yet started. \\
        Officials said attackers fired as many as 30 rockets in Friday’s assault. \\ 
        The rebel group did not immediately comment.\\
    \bottomrule
    \end{tabularx}
    \caption{A small sample of sentences in the high-likelihood region of $p(l | s_i, D)$. More examples shown in Table \ref{tab:top_fact_update_preds}.}
    \label{tab:top_fact_update_sample}
\end{table}

\paragraph{Discussion: Linguistic Cues Characterize Factual Edits. LLMs are bad at detecting these.}

Interestingly, sentence-level characteristics seem to contain much of the signal for this task: as shown in Table \ref{tab:predictive_modeling_results}, the performance barely increases by including the Full Article as context (a finding we did not observe in our tagging task, in Section \ref{sct:news-edits-schema}). To gain a deeper intuition about these sentence-level cues, we sample 100 sentences from $D_{test}^{gold}$ that have been labeled as either having a Factual Update or not (i.e. another kind of update, or no update at all). We show results in Table \ref{tab:linguistic_qualities}. We identify cues like the temporality of an event described in the sentence as important, and whether the sentence contains statistics, analysis or other kinds of news discourse \cite{van1998newsdiscourse}. Interestingly, sentences that Factual Update are more likely to contain Recent Events and Developing Events, compared with Opinion, Historical Events and Description. (See Appendix \ref{app:additional_schema_definitions} for definitions of these discourse patterns).

This would explain in part why language models underperform human reasoning in predicting updates. We find that GPT4 generally has low agreement with human annotators on these tasks, at $\kappa=.2$. Researchers have generally found that LLMs struggle with this kind of reasoning \cite{han2020econet, tan2023towards}. Recent modeling advancements might help us perform these tasks better \cite{xiong2024large}.

\textit{\ul{This prediction task is noisy: many sentences may look similar, but may or may not have had Factual Updates, due to chance. Indeed, even expert human annotators have low prediction scores.}} However, we hypothesize that data that the model is most confident about (or the high-precision region), are more uniformly predictable. We show samples of these sentences in Table \ref{tab:top_fact_update_sample}. These sentences contain many of the linguistic cues identified in \ref{tab:linguistic_qualities}. See Table \ref{tab:top_fact_update_preds} for more examples of high-probability sentences (and Table \ref{tab:bottom_lik_sents} for examples of low-probability sentences). We focus on these high-precision sentences in the next section.

\section{Part 3: Question Answering with Outdated Documents}
\label{sct:use-case}

We are ready to test whether the prediction models learned in the last section, to predict whether a sentence will have a Factual update, can help us in dynamic LLM Q\&A tasks. We set up a RealTimeQA-style task \cite{kasai2022realtime}, where an LLM is supplied by a retrieval system with potentially \textit{out-of-date} information. We would like the LLM to \textit{abstain} from answering a question if it suspects it's information might be outdated. 

Consider the scenario in Table \ref{tab:example_prompt}. As humans, we could infer that the ongoing events in the old sentence would be of relatively short time-scale. Thus, if a retriever retrieves the old sentence for the LLM, without knowledge of the new sentence, we would like the LLM to answer the question with something like: ``\textit{I do not have the most updated information and this might change quickly}''. Confidently answering without any caution as to the updating nature of events is \textit{wrong}. 

\begin{table}[t]
    \small
    \centering
    \begin{tabularx}{\linewidth}{>{\hangindent=2em}X}
    \toprule
    \textbf{Old sentence}: The White House \colorbox{red!20}{\st{is}} on lockdown after a vehicle struck a security barrier.\\
    \midrule
    \textbf{New sentence}: The White House \colorbox{green!20}{was} on lockdown \colorbox{green!20}{for about an hour} after a vehicle struck ... \\
    \midrule
    \textbf{Question}: \textit{``Can I visit the White House right now?''}\\
    \bottomrule
    \end{tabularx}
\caption{\textbf{LLM Abstention Demonstration}: In this example, the LLM only has access to the old, outdated article. We wish to probe whether LLMs can reason about the information's likelihood of being outdated and be cautious about answering this question.}
\label{tab:example_prompt}
\end{table}

\begin{table*}[t]
\small
\centering
\begin{tabular}{lrrrrrrrrrrrr}
\toprule
{} & \multicolumn{3}{c}{\textbf{No-Conflict}} & \multicolumn{3}{c}{\textbf{Maybe-Conflict}} & \multicolumn{3}{c}{\textbf{Likely-Conflict}} \\ 
{} & Micro F1 & Macro F1 & Avg. & Micro F1 & Macro F1 & Avg & Micro F1 & Macro F1 & Avg. \\
\cmidrule(lr){2-4}\cmidrule(lr){5-7}\cmidrule(lr){8-10}
No Warning & 55.9 & 35.8 & 55.9 & 8.8 & 8.1 & 8.8 & 38.8 & 28.0 & 38.8 \\
Uniform Warning & 52.9 & \textbf{49.6} & 52.9 & 90.0 & 47.4 & 90.0 & 64.7 & 54.0 & 64.7 \\
w. Update Pred. & \textbf{59.4} & 48.9 & \textbf{59.4} & \textbf{90.6} & 61.1 & \textbf{90.6} & \textbf{67.1} & \textbf{62.4} & \textbf{67.1} \\
w. Oracle Update & 57.6 & 47.7 & 57.6 & 90.0 & \textbf{63.3} & 90.0 & 66.5 & 61.1 & 66.5 \\
\bottomrule
\end{tabular}
\caption{\textbf{LLM-QA Abstention Accuracy}: we measure how often GPT4 correctly abstains from answering user-questions, based on the ground truth of whether the facts in an article updated or not. Each variant shows different information that GPT4 is given. We generate questions in three categories: No-Conflict, Maybe-Conflict, Likely-Conflict, representing how likely the answer to the question will be outdated after a factual update.}
\label{tab:use-case-results}
\end{table*}

\begin{table}[t]
    \small
    \centering
    \begin{tabular}{lrrr}
    \toprule
    {}                & \textbf{No}   & \textbf{Maybe} & \textbf{Likely} \\
    \midrule
    No Warning        & 0.0           & 0.0            & 0.0             \\
    Uniform Warning   & 30.0          & 87.1           & 98.8            \\
    w. Update Pred.   & 10.6          & 74.1           & 95.9            \\
    w. Oracle Update  & 12.4          & 75.9           & 94.1            \\
    \bottomrule
    \end{tabular}
    \caption{\textbf{Likelihood of abstaining} in the three test cases: \textbf{No} factual conflict, \textbf{Maybe} factual conflict, \textbf{Likely} factual conflict. In general, we wish to refrain only when we need to. Over-refraining is bad.}
    \label{tab:refrain}
\end{table}

\subsection{LLM-QA Experiments}

\paragraph{Experimental Design} We take pairs of sentences in the gold test set of our annotated data where an update occurred, and we ask GPT4 to ask questions based on the older sentence.

(1) \ul{No-Conflict}: 5 questions based on information in the older sentence that does \textit{NOT} update in the newer one. 

(2) \ul{Maybe-Conflict}: 5 questions based on information in the older sentence that \textit{might} update in the newer one. 

(3) \ul{Likely-Conflict}: 5 questions based on information from the older sentence \textit{likely} updates with a newer one. 
(For all prompts, see Appendix \ref{sct:prompts}). 

\paragraph{Experimental Variants} We devise the following experimental variants. Each variant take in the \textit{old sentence} and a \textit{question}, generated previously. 

(1) \ul{No Warning (Baseline \#1)}: We formulate a basic prompt to GPT4, without alerting it to any possibly outdated material. 

(2) \ul{Uniform Warning (Baseline \#2) } We warn GPT4 that some information might be outdated. The warning is the same for all questions, so GPT has to rely on its own reasoning to detect information that could be potentially outdated. 

(3) \ul{w/ Our Update Likelihood}: We give GPT4 predictions from our Factual Update model, binned into ``low'', ``medium'', ``high'' update likelihood.  (We use the highest-scoring LED variation). 

(4) \ul{w/ Oracle Update}: We give GPT4 gold labels that a fact-update \textit{did} or \textit{did NOT} occur. 
This is designed to give us an upper bound on abstention.

\paragraph{Abstention Rate Evaluations} We evaluate performance of each prompting strategy using a GPT4-based evaluation. We ask GPT4: (1) Is this question answerable given the information in the old sentence? (2) Is the answer consistent with the information presented in the revised sentence? 

\ul{We manually label a small set of 100 questions, to verify that GPT4 can perform this task, and find high agreement $\kappa > .74$ for both questions}. If the answer to both questions is yes, the LLM should attempt to provide an answer. If either of the answers is ``no'', then we want the LLM to ABSTAIN from answering. Abstaining when it \textit{should} is a success; any other answer is a failure. We show F1 scores in Table \ref{tab:use-case-results}. Interestingly, and perhaps unexpectedly, the variant with Update Predictions does as well if not better than the variant with Oracle Updates. Perhaps the categories of the prediction score helps GPT4 better understand the task compared with the simple yes/no gold labels.

The Uniform Warning (Baseline \#2) variation has surprisingly strong performance as well, perhaps an indication that GPT4 does have some emergent abilities to detect the linguistics of outdated information. However, when we examine overall abstention rates, shown in Table \ref{tab:refrain}, we find that this baseline has a far abstention rate. Meanwhile, the variant with Update Predictions abstains at nearly the same rates as that with Oracle Updates.

\section{Discussion and Conclusion}

The ability of our prediction tags to recover near-oracle performance signals that factual edit prediction can serve a useful role in LLM Q\&A. Although we have mainly tested our results in a high-likelihood region of the problem domain as a proof of concept, we suspect that if future work improves the models trained in Section \ref{sec:predictive_modeling_data}, then we will see an increase in the ability to drive such abstentions.

We do suspect there to be an inherent upper bound in our ability to model such revision patterns. Randomness undoubtedly exists in the editing and revision process; for many factual updates where, perhaps, the ethical stakes of outdated information are lower, journalists may choose not to go back and revise. We still see such work as promising. Indeed, it is surprising that, despite low scores on the modeling components for Part 1 (Edit-Intention Tagging) and Part 2 (Factual Edit Prediction), we still observe useful downstream applications in Part 3. The linguistic insights we are observe concord with human intuition, and identify known shortcomings of current language models.

Thus, we hope more broadly that the taxonomy introduced in \textit{NewsEdits 2.0} has many rich directions for yielding linguistic insights and better benchmarks. We hope in future work to revise directions around stylistic and narrative edits, both of which we believe can lead to better tools for computational journalists.

\section{Ethical Considerations}

\subsection{Dataset}

\textit{NewsEdits} is a publicly and licensed dataset under an \texttt{AGPL-3.0 License}\footnote{\url{https://opensource.org/licenses/AGPL-3.0}}, which is a strong ``CopyLeft'' license. 

Our use is within the bounds of intended use given in writing by the original dataset creators, and is within the scope of their licensing.

\subsection{Privacy}

We believe that there are no adverse privacy implications in this dataset. The dataset comprises news articles that were already published in the public domain with the expectation of widespread distribution. We did not engage in any concerted effort to assess whether information within the dataset was libelious, slanderous or otherwise unprotected speech. We instructed annotators to be aware that this was a possibility and to report to us if they saw anything, but we did not receive any reports. We discuss this more below.  

\subsection{Limitations and Risks}

The primary theoretical limitation in our work is that we did not include a robust non-Western language source. As our work builds off of NewsEdits as a primary corpora, it contains only English and French.

This work should be viewed with that important caveat. We cannot assume \textit{a priori} that all cultures necessarily follow this approach to breaking news and indeed all of the theoretical works that we cite in justifying our directions also focus on English-language newspapers. One possible risk is that some of the information contained in earlier versions of news articles was updated or removed for the express purpose that it was potentially unprotected speech: libel, slander, etc. Instances of First Amendment lawsuits where the plaintiff was successful in challenging content are rare in the U.S. We are not as familiar with the guidelines of protected speech in other countries.

We echo the risk of the original \textit{NewsEdits} authors: another risk we see is the misuse of this work on edits for the purpose of disparaging and denigrating media outlets. Many news tracker websites have been used for good purposes (e.g. holding newspapers accountable for when they make stylistic edits or try to update without giving notice). But we live in a political environment that is often hostile to the core democracy-preserving role of the media. We focus on fact-based updates and hope that this resource is not used to unnecessarily find fault with media outlets.  

\subsection{Computational Resources}

The experiments in our paper require computational resources. Our models run on a single 30GB NVIDIA V100 GPU or on one A40 GPU, along with storage and CPU capabilities provided by our campus. While our experiments do not need to leverage model or data parallelism, we still recognize that not all researchers have access to this resource level.

We use Huggingface models for our predictive tasks, and we will release the code of all the custom architectures that we construct. Our models do not exceed 300 million parameters.

\subsection{Annotators}

We recruited annotators from professional journalism networks like the NICAR listserve, which we mention in the main body of the paper. All the annotators consented to annotate as part of the experiment, and were paid \$1 per task, above the highest minimum wage in the U.S. Of our 11 annotators, all were based in large U.S. cities. 8  identify as white, 1 as Asian, 1 as Latinx and 1 as black. 8 annotators identify as male and 3 as female. This data collection process is covered under a university IRB. We do not publish personal details about the annotations, and their interviews were given with consent and full awareness that they would be published in full.

\subsection{References}

\bibliography{custom}

\clearpage
\appendix

\section{Additional EDA}
\label{app:eda}

\begin{table}[t]
    \centering
    \small
\begin{tabular}{lrrr}
\toprule
{} & Fact & Style & Narrative \\
\midrule
Disaster & {\cellcolor[HTML]{023858}} \color[HTML]{F1F1F1} 6.4 & {\cellcolor[HTML]{FFF7FB}} \color[HTML]{000000} 43.4 & {\cellcolor[HTML]{023858}} \color[HTML]{F1F1F1} 50.0 \\
Elections & {\cellcolor[HTML]{1379B5}} \color[HTML]{F1F1F1} 5.1 & {\cellcolor[HTML]{B4C4DF}} \color[HTML]{000000} 47.9 & {\cellcolor[HTML]{2D8ABD}} \color[HTML]{F1F1F1} 46.9 \\
Environment & {\cellcolor[HTML]{FFF7FB}} \color[HTML]{000000} 1.9 & {\cellcolor[HTML]{023858}} \color[HTML]{F1F1F1} 56.8 & {\cellcolor[HTML]{FFF7FB}} \color[HTML]{000000} 41.2 \\
Labor & {\cellcolor[HTML]{FCF4FA}} \color[HTML]{000000} 2.0 & {\cellcolor[HTML]{83AFD3}} \color[HTML]{F1F1F1} 49.6 & {\cellcolor[HTML]{0568A3}} \color[HTML]{F1F1F1} 48.2 \\
Other & {\cellcolor[HTML]{9CB9D9}} \color[HTML]{000000} 3.7 & {\cellcolor[HTML]{5EA0CA}} \color[HTML]{F1F1F1} 50.7 & {\cellcolor[HTML]{78ABD0}} \color[HTML]{F1F1F1} 45.5 \\
Safety & {\cellcolor[HTML]{3790C0}} \color[HTML]{F1F1F1} 4.7 & {\cellcolor[HTML]{D2D3E7}} \color[HTML]{000000} 46.6 & {\cellcolor[HTML]{046096}} \color[HTML]{F1F1F1} 48.6 \\
\bottomrule
\end{tabular}
    \caption{Distribution over update-types, across social-interest categories \cite{spangher2023identifying}.}
    \label{tab:topic_classifiation_1}
\end{table}

We show the following different analyses to support the findings in the main body. 

Table \ref{tab:topic_classifiation_1} shows the kinds of edits in 6 different categories of news determined ``socially beneficial'', by \cite{spangher2023identifying}\footnote{To group news articles in these categories, we use a classifier released by the authors}. As can be seen, even though Factual updates are rarer overall in sentence-level updates, they are more represented in Disaster and Safety categories.

\begin{figure*}[t]
    \centering
    \includegraphics[width=\linewidth, trim={0 0 5cm 0},clip]{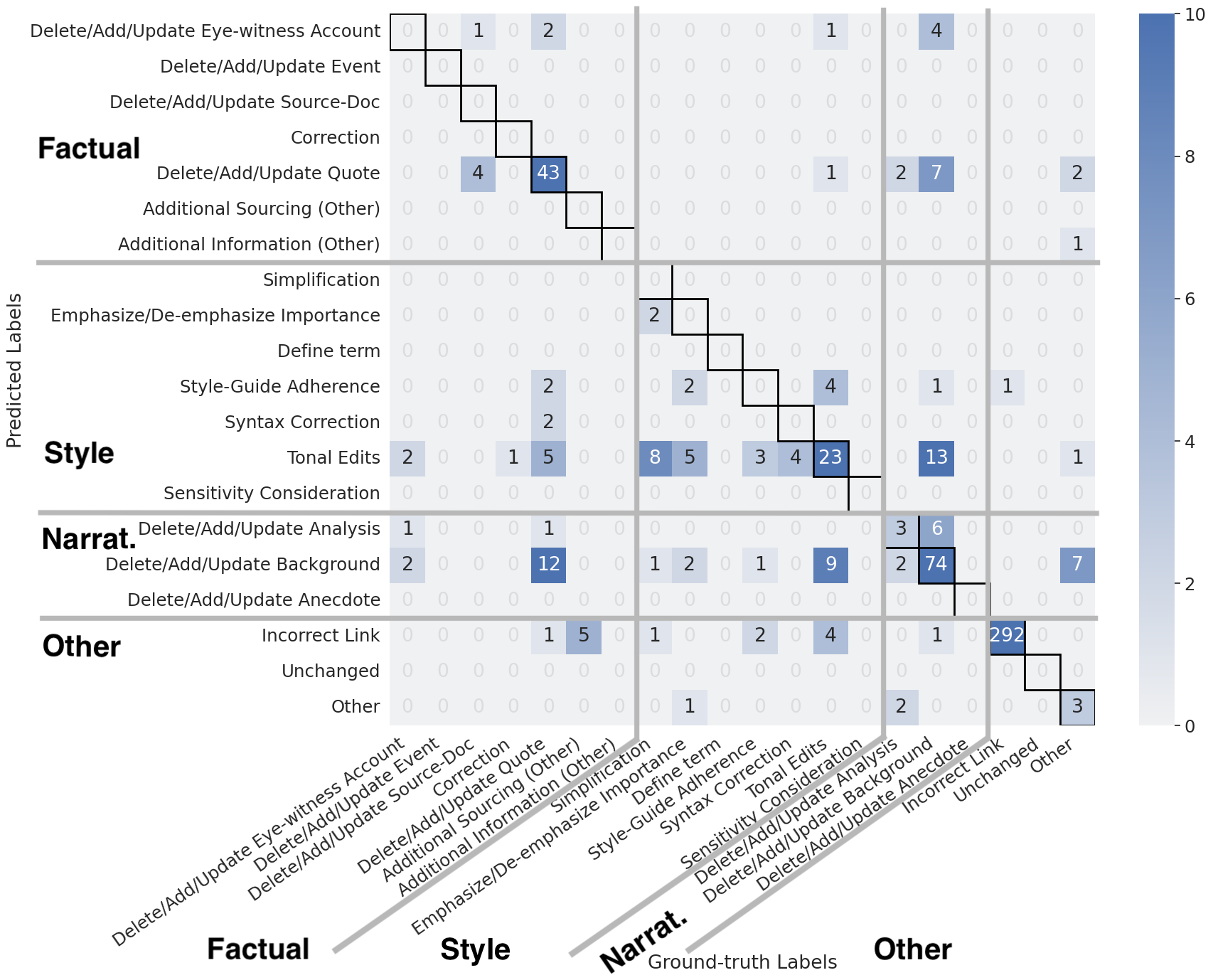}
    \caption{Fine-grained confusion matrix for the LED model trained with Discourse and Argumentation features.}
    \label{fig:confusion_matrix}
\end{figure*}

In Figure \ref{fig:confusion_matrix}, we perform an error analysis on our best-performing ensemble model, which includes tags from Argumentation and Discourse. We inspect the categories we are most likely to get wrong. As can be seen, our fine-grained accuracy is actually quite low, indicating the value of future work, perhaps collecting more training data or employing LLMs to label more silver-standard data. Many categories on the diagonal have 0 labels, both because many categories are low-count categories (e.g. ``Define Term'', which does not have \textit{any} gold-truth labels in the test set), as well as that more dominant categories capture many of the predictions (e.g. ``Tonal Edits``).

\begin{figure}[t]
    \centering
    \includegraphics[width=.8\linewidth]{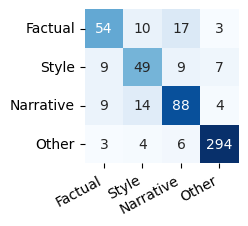}
    \caption{Coarse-grained confusion matrix for the LED model trained with Discourse and Argumentation features.}
    \label{fig:coarse-grained-confusion-matrix}
\end{figure}

However, the problem is slightly less sever on the coarse-grained level, shown in Figure \ref{fig:coarse-grained-confusion-matrix}. By comparing these two categories, we can see that many of the errors we observed are on the fine-grained level are within the same coarse-grained category. We suspect that to raise accuracy for fine-grained labels further, we need further experimentation is needed. Perhaps we can experiment with approaches involving more specific fine-grained models or with data augmentation.

\subsection{Further details about high-precision sentences}

\begin{figure}[t]
	\centering
	\includegraphics[width=.7\linewidth]{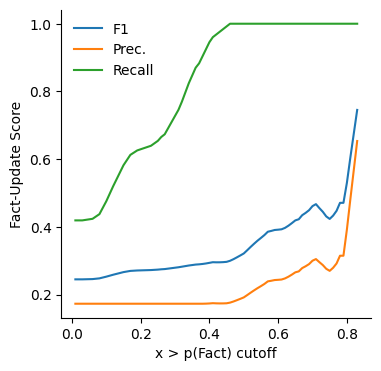}
	\caption{Performance of Fact-update model increases as we increasingly focus on a pool of documents that are categorized as high-likelihood under the top-performing LED model (in Table \ref{tab:led_results}). In other words, the model truly shines in the high-precision, high-probability realm.}
	\label{fig:high-prob}
\end{figure}

Figure \ref{fig:high-prob} shows more details of our exploration into the predictability of higher-precision fact-update sentences: as we restrict the pool of documents, we increase the performance. 

\subsection{Technical Improvements over \textit{NewsEdits} Edit-Action Algorithm}
\label{sct:technical_improvements}

\newcite{spangher2022newsedits} identified ``edit-actions'', or ``syntactic'' edits in article revision histories (i.e. sentence additions, deletions and updates), which requires them to match sentences across article versions. They report a 89.5 F1 efficacy at matching sentences, a significantly higher rate than we might expect for lexical matching. We examined \textit{NewsEdits}'s sentence matches and found that a large source of errors stem from poor sentence boundary detection (SBD). Poor SBD creates an abundance of sentence stubs, which often over-match across revisions. We reprocessed the dataset from scratch using spaCy\footnote{\url{https://spacy.io/}, specifically, the \texttt{en\_core\_web\_lg} model.} instead of SparkNLP for SBD\footnote{\url{https://sparknlp.org/api/com/johnsnowlabs/nlp/annotators/sbd/pragmatic/SentenceDetector.html}}, which we qualitatively observe to be better. For word-matching, we use \texttt{albert-xxlarge-v2}\footnote{\url{https://huggingface.co/albert/albert-xxlarge-v2}}'s embeddings \cite{lan2019albert} instead of \texttt{TinyBert} \cite{jiao2019tinybert}. These steps, we find, increase our linking accuracy to 95 F1-score. We reprocess and re-release \textit{NewsEdits}. In addition, we release a suite of visualization tools, based on D3\footnote{\url{https://d3js.org/}} to enable further exploration of the corpus. See Appendix \ref{app:annotation_interface} for an example.

\section{Details of the LED Model}
\label{apx:led_details}
In this section, we describe the specifications of the LED model described in \Cref{sct:revision-pair-modeling}.

\subsection{Input Template}
The input to the LED model is shown below:

\begin{quoting}
    \texttt{Predict the edit intention from version 1 to version 2.} \\
     \texttt{Version 1: \textbf{SOURCE\_SENTENCE}} \\
    \texttt{Version 2: \textbf{TARGET\_SENTENCE}} \\
    \texttt{Version 1 Document: \textbf{SOURCE\_DOCUMENT}} \looseness=-1\\
    \texttt{Version 2 Document: \textbf{TARGET\_DOCUMENT}}
\end{quoting}

Here, \texttt{\textbf{SOURCE\_DOCUMENT}} ($D$) and \texttt{\textbf{TARGET\_DOCUMENT}} ($D'$) refer to the newer and older articles, while \texttt{\textbf{SOURCE\_SENTENCE}} ($s_i$) and \texttt{\textbf{TARGET\_SENTENCE}} ($s'_j$) represent a sentence with these articles.

\begin{table*}[t]
    \centering
    \begin{tabular}{lrrr}
    \toprule
    {} & \textbf{Addition} & \textbf{Deletion} & \textbf{Edit} \\
    \midrule
    Add/Delete/Update Background & 806909 & 329652 & 411025 \\
    Add/Delete/Update Quote & 303451 & 17995 & 46300 \\
    Incorrect Link & 191022 & 125362 & 237437 \\
    Other (Please Specify) & 84646 & 66929 & 65077 \\
    Add/Delete/Update Event Reference & 37409 & 3645 & 56098 \\
    Add/Delete/Update Analysis & 33426 & 390 & 268 \\
    Add/Delete/Update Eye-witness account & 9772 & 0 & 3 \\
    Add/Delete/Update Source-Document & 6639 & 2 & 28 \\
    Add/Delete/Update Information (Other) & 1058 & 13 & 3 \\
    Additional Sourcing & 573 & 15 & 29 \\
    Tonal Edits & 102 & 6000 & 616514 \\
    Emphasize/De-emphasize Importance & 1 & 32 & 1076 \\
    Syntax Correction & 1 & 2 & 21729 \\
    Emphasize/De-emphasize a Point & 0 & 53 & 1668 \\
    Simplification & 0 & 0 & 3 \\
    Style-Guide Edits & 0 & 1 & 3253 \\
    Correction & 0 & 1 & 47 \\
    \bottomrule
\end{tabular}
\caption{Counts of fine-grained semantic edit types, broken out by syntactic categories}
\end{table*}

\subsection{Additional Schema}
\label{app:additional_schema_definitions}

\paragraph{NLI}

We use textual entailment from \cite{dagan2005pascal}, which consists of \textit{Entail}, \textit{Contradict} and \textit{Neutral}. These categories indicate whether two pieces of information refute each other, complement each other, or are neutral. We use a trained model by \cite{nie-etal-2020-adversarial}, which is an adversarially-trained Albert-xxlarge model, to label pairs of sentences (one from the old version, one from the new version).

\begin{figure}[t]
    \centering
    \includegraphics[width=\linewidth]{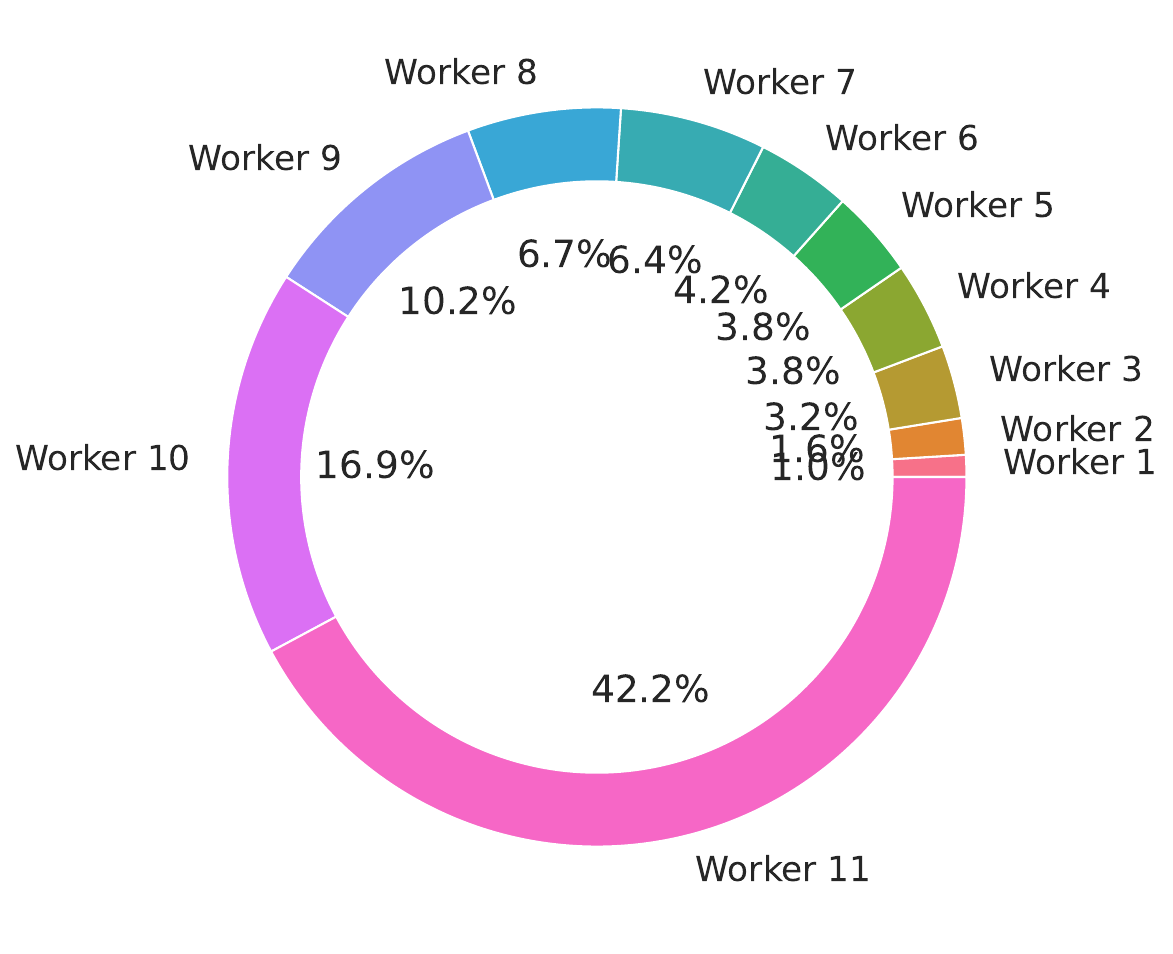}
    \caption{The portion of annotation tasks assigned to each worker.}
    \label{fig:worker_task_distribution}
\end{figure}

\paragraph{Event Detection}

As described by \newcite{doddington2004automatic} in the coding guidelines for the ACE-2005 dataset, ``\textit{An Event is a specific occurrence involving participants. An Event is something that happens. An Event can frequently be described as a change of state.}'' Several datasets exist which label events in text, like ACE-2005, and a wide body of research has since emerged to model and detect events in text. Such models detect \textit{triggers} (i.e. mostly verb-forms that signal the presence of an event); \textit{types} (i.e. broad taxonomies that events fall into) and \textit{arguments} (i.e. people, places or other lexical units associated with the occurrence of the event which further define it).

We use a model by \cite{hsu2021degree}, designed to detect events in a wide variety of settings. We only consider whether an event trigger exists in a sentence, as a binary variable (0=no trigger exists, 1=trigger exists). Our theory is that this can help with tags like ``Delete/Add/Update Event''.

\paragraph{Argumentation}

Defined in \newcite{alkhatib2016b}, \textit{Argumentation} is a type of discourse schema that defines what kinds of evidence the writer marshalls to make their point. Authors define the following categories: \textit{Anecdote}, \textit{Assumption}, \textit{Common Ground}, \textit{Statistics}, \textit{Testimony}, \textit{Other}. They primarily study news editorials (i.e. opinion pieces), where they assume they have the most different kinds of argumentation categories. \newcite{spangher2021multitask} and \newcite{spangher2023explaining} show that these models can generally be applied helpfully across a broader news domain. We include them in the present study to capture aspects like ``Anecdote'' that capture framing aspects of journalistic writing.

\paragraph{Quote}

Quote-detection is a long-standing task, usually involving detecting the presence of direct or indirect quotes \cite{pareti2013automatically}. We use the broad definition of a ``quote'' as ``information derived from any source external to the news article and the journalist's own thoughts'', as defined in \newcite{spangher2023identifying}. Authors developed and released models for detecting when sentences had information that could be attributable to a named or unnamed source in the news article. We use these models to apply a simple binary indicator for whether or not the sentence contained a quote (1=sentence contains a quote, 0=it does not). We include this under the hypothesis that it can help us improve our detection in categories like ``Delete/Add/Update Quote''.

\paragraph{News Discourse}

The News Discourse schema, as defined by \newcite{van1998newsdiscourse} views news stories as a sequence of structural elements, each serving a different narrative role. As implemented separately by \cite{choubey2020discourse}, \cite{yarlott2018identifying} and \cite{spangher2021multitask}, the news discourse schema has undergone some modifications since \newcite{van1998newsdiscourse}'s original formulation, most notably to include current theories on event detection. It includes the following elements: \textit{Main Event}, \textit{Consequence}, \textit{Previous Event},  \textit{Current Context}, \textit{Evaluation}, \textit{Expectation}, \textit{Historical Events}, \textit{Anecdotal Event}. We believed that, since much of our edit schema was inspired by notions of narration, like ``Delete/Add/Update Background'', we could get signal from this schema.

\section{Annotation Details}
In this section, we provide details of the annotation process, such as annotation guidelines and task allocation.

\subsection{Annotation Guidelines}
\label{apx:annotation_guidelinnes}
To complete the task, look at each sentence: if it's been added, updated, or deleted between drafts, try to determine based on your knowledge of the journalistic editing process why this was done.

You can specify multiple intentions for each add/delete/edit operation. Please also pay attention to when sentences are moved around in a document (i.e. if that was done to emphasize or de-emphasize that sentence), and when there might be errors to how we are linking sentences.

We devised these in consultation with professional journalists. However, if you are consistently annotating edits with "Other" (i.e. we are missing something in our schema), please let us know!

\paragraph{Fact Edits:}
\begin{itemize}
    \item \textbf{Delete/Add/Update Eye-witness Account}: The writer deletes/adds/updates the contents for the events being described. This can either take the form of a quote (in which case this edit should be paired with a Quote Update), or a first-person account by the journalist.
    \item \textbf{Delete/Add/Update Event}: There is a change to some event in the world that the article covers and the article needs to be updated to reflect this. Usually, there are changes to the verbs in the article, but this can also include increased death counts, stock-market changes, etc.
    \item \textbf{Delete/Add/Update Source-Doc}: Additional written documents have been released by a government or company that warrant deletion/inclusion/update of the content of the article. For example, additional information included in an SEC filing, quarterly earnings report, IPCC report, etc.
    \item \textbf{Correction}: There are factual errors in the original version. The new version corrects the error.
    \item \textbf{Delete/Add/Update Quote}: There is an addition, editing or deletion of quotes in the article. Or, a quote from one person is swapped for a quote from another. Sometimes these updates are made with other intentions (e.g. to include a punchier quote, in which case it would also be a Preferential Edit. In these cases, please use the “+” button to add another intention dropdown.)
    \item \textbf{Additional Sourcing (Other)}: The new version includes evidence of new sources for additional information, usually added for confirmation purposes. Note that this is different from Quote Update or Document Update since Additional Sourcing doesn’t have to result in a new quote or document reference. Can simply be an indication that the journalist obtained new evidence.
    \item \textbf{Additional Information (Other)}: This edit intention is applied when the new version of the article includes details or context not present in the original version, which doesn't necessarily fall under specific updates like eye-witness accounts, event changes, document updates, or sourcing alterations. 
\end{itemize}

\paragraph{Style Edits:}
\begin{itemize}
    \item \textbf{Simplification}: educes the complexity or breadth of discussion. This edit might also remove information from the article.
    \item \textbf{Emphasize/De-emphasize Importance}: The sentence is moved up or down in the document in order to make the sentence MORE/LESS prominent, or to emphasize/de-emphasize it’s connection to the events being described in another sentence.
    \item \textbf{Define term}: The author provides meaning or differentiation to a term or concept that might be unknown to the reader. Note that this intention is DIFFERENT from the Background intention, which is more about providing context, e.g. historical or geographic context for a person, company, or place.
    \item \textbf{Style-Guide Adherence}: Edits that are made specifically to address a formal style guide (when in doubt, defer to the Associated Press style-guide). The first version violates the style guide and the revised version fixes it.
    \item \textbf{Syntax Correction}: Improve grammar, spelling, or punctuation. These are strictly to correct errors in syntax, not \textbf{Preferential Edits}. And, they need not be adhering to a formal style-guide (when a \textbf{Syntax Correction} is also adhering to a \textbf{Style Guide}, please use the “+” button to add another intention dropdown and annotate both).
    \item \textbf{Tonal Edits}: The journalist or copy-editor made the edits due to a specific personal or artistic preference. Use your intuition here: these are usually edits that introduce punch, elegance or scenery. These edits often also have the effect of some other edit intention, see the example, but cannot be fully ascribed to other aims.
    \item \textbf{Sensitivity Consideration}: The journalist rewrote the sentence because the original version is inappropriate/ may be considered insensitive.
\end{itemize}

\paragraph{Narrative Edits:}
\begin{itemize}
    \item \textbf{Delete/Add/Update Analysis}: The writer deletes/adds/updates inferences from the presented information. These can be in the form of analyses, expectations, or deeper understandings. These are usually forward-looking rather than Background information, which is usually past-looking.
    \item \textbf{Delete/Add/Update Background}: Delete/add/update contextualizing information to the article to help readers understand the history, geography or significance of a term, personal, place or company. Note that contextualizing information is not analysis, expectations, or projections, which would fall into the Analysis intention category.
    \item \textbf{Delete/Add/Update Anecdote}: The writer deletes, adds, or updates a brief, revealing account of a person or event. This can be a personal story, a particular incident, or a narrative snippet that exemplifies a point or adds a humanizing or illustrative dimension to the news piece. These anecdotes may serve to engage the reader's interest, illuminate a fact, or provide a real-world example of abstract concepts.
\end{itemize}

\paragraph{Others:}
\begin{itemize}
    \item \textbf{Incorrect Link}: This refers to an error in our original linking of sentences. We have linked two sentences that should NOT be linked. This only pertains to `Edit`ed or `Unchanged` sentences. Sentences should not be linked if they are entirely unrelated — they have substantially different syntax, intent, and purpose — and, by error, our algorithm said they were. If you identify an \textbf{Incorrect Link} AND there are more than one links, please specify (A) the index of the sentence in the other version that it should NOT be linked to via the dropdown (B) any other intention ascribed to this pair (i.e. Fact Deletion).
\end{itemize}

\subsection{Annotation Interface}
\label{app:annotation_interface}
\begin{figure*}[t]
    \centering
    \includegraphics[width=\linewidth]{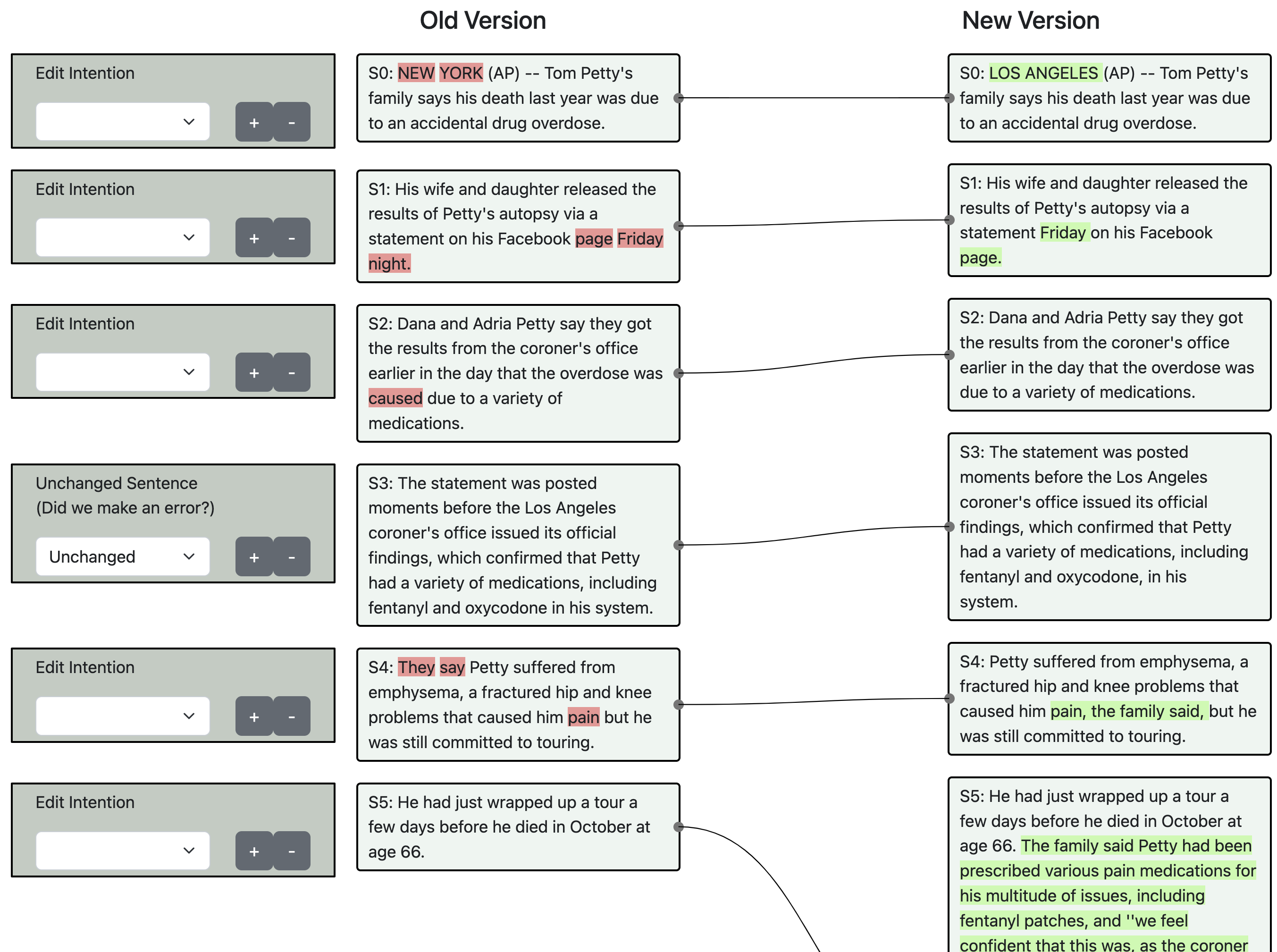}
    \caption{The interface for annotating edit intentions.}
    \label{fig:annotation_interface}
\end{figure*}

\Cref{fig:annotation_interface} shows the annotation interface for our task. Users are shown pairs of sentences, as identified in NewsEdits \cite{spangher2022newsedits} and have the option to annotate edits, additions and deletions with different edit intentions. Additionally, users can annotate when the links are incorrect.

\subsection{Annotation Task Distribution}

We asked prospective applicants to describe their journalism experience, and selected this pool based on those having one or more year of professional editing experience. Then, we asked them to label revised sentences in five news articles, which we checked. We recruited 11 annotators who scored above 90\% on these tests. 

In \Cref{fig:worker_task_distribution}, we show the portion of annotation tasks assigned to each worker. As can be seen, we have a broad mix of users. Worker 11 is a professional journalist we worked most often with, and annotated a plurality of the tasks.

\section{Prompts for Use-Case}
\label{sct:prompts}

\subsection{Question-Asking Prompts}

\subsubsection{No-Conflict}
\paragraph{Prompt Outline}

\begin{quoting}
\texttt{I will give you a sentence and you will give me 5 different questions. It should be directly answerable by the sentence.}\\
\texttt{Here are some examples:}\\
\texttt{Example 1: \textbf{EXAMPLE}}\\
\texttt{Example 2: \textbf{EXAMPLE}}\\
\texttt{Example 3: \textbf{EXAMPLE}}\\
\texttt{Ok, now it's your turn.}\\
\texttt{Here is a sentence: \textbf{SENTENCE}}
\texttt{Ask 5 different questions, output in a list. Don't say anything else.}
\end{quoting}

\paragraph{Examples}

\textit{sentence}: "WASHINGTON (AP) -- The White House is on lockdown after a passenger vehicle struck a security barrier."
\textit{question}: "What did the vehicle strike?"

\textit{sentence}: "The death count from the 42nd street bombing is 49 injured, 2 killed so far."
\textit{question}: "Where did the bombing take place?"

\textit{sentence}: "The construction work left the bridge badly damaged and unsafe for passengers and is expected to remain so for days."
\textit{question}: "What kind of work was being done?"

\subsubsection{Maybe-Conflict}
\paragraph{Prompt Outline}

\begin{quoting}
\texttt{I will give you a sentence and you will give me an answer. It should be timely and related to the facts in the sentence. 
It should be a question that could go stale, especially for ongoing events, or facts like death counts that might update.} \\
\texttt{Here are some examples:}\\
\texttt{Example 1: \textbf{EXAMPLE}}\\
\texttt{Example 2: \textbf{EXAMPLE}}\\
\texttt{Example 3: \textbf{EXAMPLE}}\\
\texttt{Ok, now it's your turn.}\\
\texttt{Here is a sentence: \textbf{SENTENCE}}
\texttt{Ask 5 different questions, output in a list. Don't say anything else.}
\end{quoting}

\paragraph{Examples}
\textit{sentence}: "WASHINGTON (AP) -- The White House is on lockdown after a passenger vehicle struck a security barrier."
\textit{question}: "Is the White House currently in lockdown -- if I visit, will I get turned away?"

\textit{sentence}: "The death count from the street bombing is 49 injured, 2 killed so far."
question: "How many people have died so far?"

\textit{sentence}: "The construction work left the bridge badly damaged and unsafe for passengers and is expected to remain so for days."
\textit{question}: "What route should I take? The bridge is the quickest way to work."

\subsubsection{Likely Conflict}

\paragraph{Prompt Outline}
\begin{quoting}
\texttt{I will give you two sentences from an updating news article and you will give me 5 different questions.
They should ideally focus on information that changes in between the sentences. So, if someone were to just look at the old sentence
and you asked them your question, they would get it wrong.}\\\
\texttt{Ok, now it's your turn.}
\texttt{Here is the old sentence: \textbf{OLD\_SENTENCE}}
\texttt{Here is the new sentence: \textbf{NEW\_SENTENCE}}
\texttt{Ask 5 different questions, output in a list. Don't say anything else.}
\end{quoting}

\paragraph{Examples}

\textit{old sentence}: "WASHINGTON (AP) -- The White House is on lockdown after a passenger vehicle struck a security barrier."
\textit{new sentence}: 'WASHINGTON (AP) -- The White House was on lockdown for about an hour Friday after a passenger vehicle struck a security barrier.'
\textit{question}: "Is the White House currently in lockdown -- if I visit, will I get turned away?"

\textit{old sentence}: "ISTANBUL (AP) -- An earthquake with a preliminary magnitude of 6.2 shook western Turkey and the Greek island of Lesbos Monday, scaring residents and damaging buildings."
\textit{new sentence}: "ISTANBUL (AP) -- An earthquake with a preliminary magnitude of 6.2 shook western Turkey and the Greek island of Lesbos on Monday, injuring at least 10 people and damaging buildings, authorities said."
\textit{question}: "Was anyone injured?"

\textit{old sentence}: "Turkey's emergency management agency said there were no reports of casualties in the country."
\textit{new sentence}: "Turkey's emergency management agency said there were no reports of casualties and has dispatched emergency and health teams, and 240 family tents to the area as a precaution."
\textit{question}: "Is the Turkish emergency management doing anything as a precaution?"

\subsection{Question Answering Prompts}

\subsubsection{Experimental Prompt}
\begin{quoting}
\texttt{You are a helpful assistant who answers questions based on this news information:}\\
\texttt{\textbf{NEWS\_ARTICLE\_SENTENCE}}\\ \\
\texttt{We give this a \textbf{HIGH/MEDIUM/LOW} chance of there being a fact update in this sentence. That might mean some new information could make some of the information in this sentence outdated.
    The user will ask a question. Answer cautiously and do not give the user wrong/outdated information.
    If the user's question looks like it will still be relevant even if the facts change, answer it directly.
    If the user's question looks like it will be outdated, say "I don't have the most up-to-date information" and that's it. Say nothing else. Do NOT say "I don't have the most up-to-date information" AND something else.}\\ \\ 
\texttt{    Keep our estimate in mind.}\\
\end{quoting}

\subsubsection{Baseline 1} 
\begin{quoting}
\texttt{    You are a helpful assistant who answers questions based on this news information:}\\
\texttt{\textbf{NEWS\_ARTICLE\_SENTENCE}}\\ \\
\texttt{    Try to directly answer the users question and say nothing else.}
\end{quoting}

\subsubsection{Baseline 2}
\begin{quoting}
\texttt{You are a helpful assistant who answers questions based on this news information:}\\
\texttt{\textbf{NEWS\_ARTICLE\_SENTENCE}}\\ \\ 

\texttt{    This sentence might go out of date.
Answer cautiously and do not give the user wrong/outdated information.
If the user's question looks like it will still be relevant even if the facts change, answer it directly.
If the user's question looks like it will be outdated, say "I don't have the most up-to-date information" and that's it. }\\ \\
\texttt{    Say nothing else. Do NOT say "I don't have the most up-to-date information" AND something else.}
\end{quoting}

\subsubsection{Oracle} 
\begin{quoting}
\texttt{You are a helpful assistant who answers questions based on this news information:}\\
\texttt{\textbf{NEWS\_ARTICLE\_SENTENCE}}\\

\texttt{    This sentence \textbf{DOES / DOES NOT} have a major fact update. That might mean some new information, updating information.
Answer cautiously and do not give the user wrong/outdated information.
If the user's question looks like it will still be relevant even if the facts change, answer it directly.
If the user's question looks like it will be outdated, say "I don't have the most up-to-date information" and that's it. }\\ \\
\texttt{    Say nothing else. Do NOT say "I don't have the most up-to-date information" AND something else.}
\end{quoting}

\subsection{Evaluation Prompts}

\begin{quoting}
\texttt{You are a helpful assistant. You will be shown an old sentence, a revised sentence, and a user-question.
you will answer the following 2 questions:}\\
\texttt{1. Is this question answerable given JUST the old sentence?}\\
\texttt{    Answer with "yes" or "no". Do not answer anything else.
    If the answer to 1 was yes, then proceed to the second question, otherwise respond to question 2 with n/a}\\
\texttt{2. Does the question ask about something that is factually consistent with the information presented in the revised sentence?}\\
\texttt{    Answer with "yes", "no" or "n/a." Do not answer with anything else.}
\end{quoting}

\begin{table*}[t]
    \centering
    \begin{tabularx}{\linewidth}{>{\hangindent=2em}X}
    \toprule
    Top Predictions for Content Evolution Prediction, $p(l=\text{Fact Update} | s_i, D)$ \\
    \midrule
    The company takes this recommendation extremely seriously,” it said in a statement. \\
    KABUL, Afghanistan — An Afghan official says a powerful suicide bombing has targeted a U.S. military convoy near the main American Bagram Air Base north of the capital Kabul. \\
    WASHINGTON — The U.S. carried out military strikes in Iraq and Syria targeting a militia blamed for an attack that killed an American contractor, a Defense Department spokesman said Sunday. \\
    Mr. Causey, who reported his concern to authorities, was not charged in the indictment, which a grand jury returned last month, and did not immediately comment. \\
    His trial has not yet started. \\
    MEXICO CITY — A fiery freeway accident involving a bus and a tractor-trailer killed 21 people in the Mexican state of Veracruz on Wednesday, according to the authorities and local news outlets. \\
    The indictment accuses Mr. Hayes, a former congressman, of helping to route \$250,000 in bribes to the re-election campaign of Mike Causey, the insurance commissioner. \\
    No Kenyans died in the attack, Kenya’s military spokesman Paul Njuguna said Monday. \\
    Mr. Manafort, 70, will most likely be arraigned on the new charges in State Supreme Court in Manhattan later this month and held at Rikers, though his lawyers could seek to have him held at a federal jail in New York, the people with knowledge said. \\
    Officials said attackers fired as many as 30 rockets in Friday’s assault. \\
    KABUL, Afghanistan — Gunmen attacked a remembrance ceremony for a minority Shiite leader in Afghanistan’s capital on Friday, wounding at least 18 people, officials said. \\
    BEIRUT — A senior Turkish official says Turkey has captured the older sister of the slain leader of the Islamic State group in northwestern Syria, calling the arrest an intelligence “gold mine. ” \\
    Paul J. Manafort, President Trump’s former campaign chairman who is serving a federal prison sentence, is expected to be transferred as early as this week to the Rikers Island jail complex in New York City, where he will most likely be held in solitary confinement while facing state fraud charges, people with knowledge of the matter said. \\
    The watchdog, the Securities and Exchange Surveillance Commission, said Tuesday it made the recommendation to the government’s Financial Services Agency on the disclosure documents from 2014 through 2017. \\
    There are no immediate reports of casualties. \\
    It said the U.S. hit three of the militia’s sites in Iraq and two in Syria, including weapon caches and the militia’s command and control bases. \\
    The rebel group did not immediately comment. \\
    Kep provincial authorities later announced a total of five dead and 18 injured. \\
    QUETTA, Pakistan — Attackers used a remotely-controlled bomb and assault rifles to ambush a convoy of Pakistani troops assigned to protect an oil and gas facility in the country’s restive southwest, killing six soldiers and wounding four, officials said Tuesday. \\
    WASHINGTON — Senator Bernie Sanders of Vermont raised \$18.2 million over the first six weeks of his presidential bid, his campaign announced Tuesday, a display of financial strength that cements his status as one of the top fund-raisers in the sprawling Democratic field. \\
    \bottomrule
    \end{tabularx}
    \caption{Sample of the most likely fact-update sentences, as judged by our top-performing model. Top predictions reflect a combination of statistics, recent or upcoming events, and waiting for quotes.}
    \label{tab:top_fact_update_preds}
\end{table*}

\begin{table*}
    \begin{tabularx}{\linewidth}{>{\hangindent=2em}X}
    \toprule
    Lowest Predictions for Content Evolution Prediction, $p(l=\text{Fact Update} | s_i, D)$ \\
    \midrule
Sir Anthony Seldon, vice-chancellor of the University of Buckingham, said: "Cheating should be tackled and the problem should not be allowed to fester any longer. " \\
He added: "This shows the extent to which a party which had such a proud record of fighting racism has been poisoned under Jeremy Corbyn. " \\
But he said his dream of making it in the game had turned into a nightmare. “ \\
Adam Price, Plaid Cymru leader, said: "There is now no doubt that Wales should be able to hold an independence referendum. " \\
Others told how excited they had been when they were scouted by Higgins. “ \\
The former Conservative deputy prime minister said it was “complete nonsense” to suggest Brexit could be done by Christmas. “ \\
He said the QAA identified 17,000 academic offences in 2016 - but it was impossible to know how many cases had gone undetected. " \\
Nationalism leads a "false trail" in ""exactly the opposite direction", he argued, "one that pits working people against each other, based on the accident of geography". \\
He also suggested that universities should adopt "honour codes", in which students formally commit to not cheating, and also recognise the consequences facing students who are subsequently caught. \\
He added: "But my experience is, if you make that threat, you don't actually need to follow through with the dreaded milkshake tax. " \\
He said: “There’s an anger inside of me, a feeling of disgust that turns my stomach. ” \\
Damian Hinds says it is "unethical for these companies to profit from this dishonest business". \\
She added: “His plan to hold another two referendums next year – and all the chaos that will bring – will mean that his government will not have time to focus on the people’s priorities. “ \\
We would be happy to talk to the Department of Education about their concerns." ' \\
I am determined to beat the cheats who threaten the integrity of our system and am calling on online giants, such as PayPal, to block payments or end the advertisement of these services - it is their moral duty to do so," said Mr Hinds. \\
The chief executive of Action on Smoking and Health, Deborah Arnott, also warned it would be a "grave error" to move away from taxing cigarettes. " \\
Rather than just taxing people more, we should look at how effective the so-called 'sin taxes' really are, and if they actually change behaviour. " \\
He added: "How many more red lines will be laid down by sensible Labour MPs, only for the leadership to trample right over them? \\
This shows that the complaints process is a complete sham," she tweeted. " \\
Mr Hinds added that such firms are "exploiting young people and it is time to stamp them out". " \\
One said he was abused by Higgins in a gym. \\
\bottomrule
\end{tabularx}
\caption{Sample of the least likely fact-update sentences, as judged by our best-performing model. Predictions represent a combination of opinion quotes or anecdotes, projects and longer-term plans.}
\label{tab:bottom_lik_sents}
\end{table*}

\end{document}